\documentclass[runningheads]{llncs}
\usepackage{graphicx}
\usepackage{subfig}
\usepackage{multirow}
\usepackage{hyperref}
\usepackage{amsmath}
\usepackage{bbding}
\usepackage{pifont}
\usepackage{wasysym}
\usepackage{amssymb}
\usepackage{amsmath,amsfonts,amssymb}
\usepackage{graphicx}
\usepackage{setspace}
\usepackage{booktabs}
\usepackage{makecell}

\begin{document}

\title{RS-Net: Regression-Segmentation 3D CNN for Synthesis of Full Resolution Missing Brain MRI in the Presence of Tumours}
\titlerunning{RS-Net for modality synthesis}

\author{Raghav Mehta , Tal Arbel}

\authorrunning{R. Mehta, T. Arbel}

\institute{McGill University, Montreal, Quebec, Canada}

\maketitle              

\begin{abstract}
Accurate synthesis of a full 3D MR image containing tumours from available MRI (e.g. to replace an image that is currently unavailable or corrupted) would provide a clinician as well as downstream inference methods with important complementary information for disease analysis. In this paper, we present an end-to-end 3D convolution neural network that takes a set of acquired MR image sequences (e.g. T1, T2, T1ce) as input and concurrently performs (1) regression of the missing full resolution 3D MRI (e.g. FLAIR) and (2) segmentation of the tumour into subtypes (e.g. enhancement, core). The hypothesis is that this would focus the network to perform accurate synthesis in the area of the tumour. Experiments on the BraTS 2015 and 2017 datasets \cite{brats} show that: (1) the proposed method gives better performance than state-of-the art methods in terms of established global evaluation metrics (e.g. PSNR), (2) replacing real MR volumes with the synthesized MRI does not lead to significant degradation in tumour and sub-structure segmentation accuracy. The system further provides uncertainty estimates based on Monte Carlo (MC) dropout \cite{DB} for the synthesized volume at each voxel, permitting quantification of the system's confidence in the output at each location. 


\end{abstract}

\section{Introduction}
\label{sec:intro}

The presence of a variety of different Magnetic Resonance (MR) sequences (e.g.  T1, T2, Fluid Attenuated Inverse Recovery (FLAIR)) improves the analysis in the context of neurological diseases such as multiple sclerosis and brain cancers, because different sequences provide complementary information. In particular, the accuracy of detection and segmentation of lesions and tumours greatly increases should several sequences of MR be available \cite{HEMIS}, as different sequences assist in differentiating healthy tissues from focal pathologies. However, in real clinical practice, not all MR image sequences are always available for each patient for a variety of reasons, including cost or time constraints, or at times, images are available but not usable, for example due to corruption from noise or patient motion. As such, both clinical practice and automatic segmentation techniques would benefit greatly from the synthesis of one or more of the missing 3D MR image sequences based on the others provided \cite{SynthUse}. However, synthesis of full 3D brain MR image is challenging especially in the presence of pathology as different MR sequences represent pathology in a different way. 

Recently, modality synthesis has gained some attention from the medical image analysis community \cite{RF,SD,LSDN}. Several approaches have been explored, such as patch-based random forest \cite{RF} and sparse dictionary reconstruction \cite{SD}. Regression Ensembles with Patch Learning for Image Contrast Agreement (REPLICA) \cite{RF} was developed to synthesize T2-weighted MRI from T1-weighted MRI using the bagged ensemble of random forests based on nonlinear patch regression. Given the success of Convolutional Neural Networks (CNNs) \cite{CIC} and Generative Adversarial Networks (GANs) \cite{pix2pix} for image-to-image translation in the field of computer vision, several recent 2D CNN \cite{MRMR,LSDN} and 2D GANs \cite{MRCT} have been developed for modality synthesis in the context of medical imaging, showing promising results for synthesis of healthy subject MRI. A patch-based Location Sensitive Deep Network (LSDN) \cite{LSDN} was developed to combine intensity and spatial information for synthesizing T2 MRI from T1 MRI and vice versa. A 2D CNN model was developed to generate 2D synthesized images with missing input MRI \cite{MRMR}. Quantitative analysis showed superior performance over competing methods based on global image metrics (PSNR and SSIM). However, the performance of the method in the area of focal pathology was not examined. 

In this paper, an end-to-end 3D CNN is developed that takes as input a set of acquired MRI sequences of patients with tumours and simultaneously performs (1) regression to generate a full resolution missing 3D MR modality and (2) segmentation of the brain tumour into subtypes. The hypothesis is that by performing regression and segmentation concurrently, the network should produce full-resolution, high quality 3D MR images, particularly the area of the tumour. The network is trained and tested on the MICCAI 2015 and 2017 BraTS datasets \cite{brats}. In the first set of experiments, the framework is evaluated against state-of-the-art synthesis methods \cite{RF,LSDN,MRMR} based on global image metrics used in previous work \cite{MRMR}, where it is shown to slightly outperform all reported results. The second set of experiments evaluate the synthesis quality at pathological locations, by examining its performance on subsequent independent downstream tasks, namely tumour segmentation. Results show that real MR images can be swapped with the generated synthesized T1, T2, and FLAIR MR images with minimal loss in segmentation performance. The network also quantifies the uncertainty of the regressed synthetic volumes through Monte Carlo dropout~\cite{DB}. This permits the confidence in the synthesis results to be conveyed to radiologists and clinicians and to automatic downstream methods that would use the synthesized volumes as inputs. 

\section{Regression-Segmentation CNN Architecture}

\begin{figure*}[!t]
\centering
\includegraphics[width=0.97\textwidth]{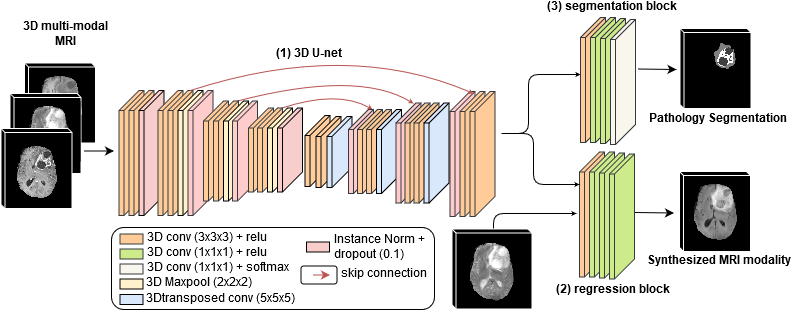}
\center \caption{Proposed Regression-Segmentation CNN architecture (RS-Net): (1) A 3D U-net, (2) Regression and (3) Segmentation convolution blocks. The model takes as input several full 3D MR image sequences, synthesizes the missing 3D MRI, while concurrently generating the multi-class segmentation of the tumour into sub-types.}
\label{fig:SynthSegCNN}
\end{figure*}

A flowchart of the proposed Regression-Segmentation CNN architecture (\textbf{RS-Net}) can be seen in Figure \ref{fig:SynthSegCNN}. The network consists of three main components: (1) a modified 3D U-net \cite{3DUnet}, (2) regression convolution block for synthesizing image sequence, and (3) segmentation convolution block for multi-class tumour segmentation. RS-Net takes as input full 3D volumes of all available sequences of a patient. The U-net generates an intermediate latent representation of the inputs which is provided to the regression and the segmentation convolution blocks. These then generate synthesis of the missing 3D MR image sequences and multi-class segmentation of tumours into sub-types, at the same resolution. The U-net learns latent representation which is common to both tumour segmentation and synthesis, with focus on high accuracy in the area containing tumour structures. In addition to the U-net output, the regression block is also provided with one of the input MRIs, which will provide necessary brain MR context to the regression block. The architecture details are now described.

The 3D U-net is similar to the one proposed in \cite{3DUnet}, with some modifications. The U-net consists of 4 resolution steps for both encoder and decoder paths. At the start, we use 2 consecutive 3D convolutions of size 3x3x3 with \textit{k} filters, where \textit{k} denotes the user-defined initial number of convolution filters. Each step in the encoder path consists of 2 3D convolutions of size 3x3x3 with $k * 2^n$ filters, where \textit{n} denotes the U-net resolution step. This is followed by maxpooling of size 2x2x2. At the end of each encoder step, instance normalization \cite{InstanceNorm} is applied, followed by dropout \cite{DropOut} with $0.1$ probability. In the decoder path at each step, 3D transposed convolution of size 5x5x5 is applied, with 2x2x2 stride and $k * 2^n$ filters for the upsampling task. The output of the transposed convolution is concatenated with the corresponding output of the encoder path. This is, once again, followed by instance normalization and Dropout with $0.1$ probability. Finally, 2 3D convolution of size 3x3x3 with $k * 2^n$ filters are applied. Rectified linear unit is chosen as a non-linearity function for every convolution layer.

Each of the segmentation and regression blocks contain 4 convolution layers. The first convolution layer is of size 3x3x3, and the rest are of size 1x1x1. The first three convolution layers have $k*4$, $k*2$ and $k$ filters. In the regression block, the last layer has just 1 filter, while, for the segmentation block, there are $C$ filters in the last layer, where $C$ denotes the total number of classes for the segmentation task. 

Weighted Mean Squared Error (MSE) loss is used for the synthesis task, and weighted Categorical Cross Entropy (CCE) loss for segmentation. Here, the weights are defined such that the weight increases whenever there are fewer voxels in a particular class.  

\begin{equation}
w_n^i = w_l*y_n^i   \qquad\text{where,} \      w_l = (\frac{\sum_{k=0}^{k=C} m_k}{m_l}) * r^{ep} + 1, 
\end{equation}

\noindent where, $w_n^i$ and $w_l$ denote the weight for voxel $n$ of volume $i$ and the weight of class $l$. $m_l$ is total number of voxels of $l^{th}$ class in the training dataset. $w_l$ are decayed over each epoch $ep$ with a rate of $r \in [0,1] $. It should be noted that $w_l$ converges to $1$ as $ep$ becomes large. The final loss function for the network, $L^i$, (for volume $i$) is a weighted combination of both of these loss functions: 

\vspace*{-2.5mm}
\begin{equation}
L^i = \lambda_1(MSE^i) + \lambda_2(CCE^i).
\label{loss_eq}
\end{equation}

Given the challenges associated with regressing a synthesized volume, errors are bound to exist. As such, deterministic outputs present dangers to subsequent clinical decisions as well as to downstream automatic methods that make use of the results. In this work, the network output is augmented with uncertainty estimates based on Monte Carlo dropout \cite{DB}. During testing, $N$ Monte Carlo (MC) samples of the output are acquired by passing each set of input volumes $N$ times through the network to predict $N$ different synthesized output MR volumes with probability of randomly dropping any neuron of the network equal to the dropout rate. Uncertainty in the synthesized volume, during testing, is estimated based on the variance of the MC samples at every voxel.

\section{Experiments and Results}
\label{sec:exptsresults}

We now evaluate the performance of the RS-Net using two sets of experiments. In the first set of experiments, we compare the quality of the synthesized volume generated by RS-Net against other methods \cite{MRMR,RF,LSDN} using PSNR and SSIM on 2015 MICCAI BraTS dataset \cite{brats}. In the second set of experiments, we evaluate the quality of the synthesized volumes in a downstream task of tumor segmentation on 2017 MICCAI BraTS datasets \cite{brats}. 

RS-Net uses 4 initial convolutional filters and 4 steps for U-net encoder and decoder paths.This results in a network with a total of 674455 learnable parameters. Values of $\lambda_1$ and $\lambda_2$ in the loss function (Eq. \ref{loss_eq}), to combine CCE and MSE, were fixed to $1.0$ and $0.1$ respectively based on experimentation evidence. The networks were trained on a NVIDIA Titan Xp GPU for 240 epochs. Approximate training time was 3 days. The networks were trained with batch size of 1, using Adam optimizer \cite{adam} with the following hyperparameters: learning rate $=0.0002$, $\beta_1 =0.9$, $\beta_2=0.999$ and $\epsilon=10^{-08}$. During testing time, a total of 20 samples of the output were generated to estimate the uncertainty in the synthesized volumes.

\subsection{Comparison of RS-Net synthesis results against other methods}
\label{OtheMethods}
In order to compare the quality of the synthesized volumes produced by RS-Net against other state-of-the-art methods, namely REPLICA \cite{RF}, LSDN \cite{LSDN}, and 2D CNN \cite{MRMR}, we train two different RS-Nets for T2 and FLAIR synthesis from T1 MRI, as done by Chartsias et al. \cite{MRMR}. We use the evaluation metrics, SSIM \cite{ssim} and PSNR, defined in \cite{MRMR}, to evaluate the quality of the synthesized volumes. 

Given a ground-truth volume $X$ and its corresponding synthesized volume $\hat{X}$, SSIM is computed as $SSIM(X,\hat{X}) = \frac{(2{\mu}_{X}{\mu}_{\hat{X}} + c_1) (2 {\sigma}_{X\hat{X}} + c_2)}{({\mu}_{X}^{2} + {\mu}_{\hat{X}}^{2} + c_1) ({\sigma}_{X}^{2} + {\sigma}_{\hat{X}}^{2} + c_1)}$, where ${\mu}_{X}$ and ${\sigma}_{X}^{2}$ are mean and variance of volume $X$ and ${\sigma}_{X\hat{X}}$ is the covariance between $X$ and $\hat{X}$. PSNR is computed as $10 \log_{10} (\frac{{MAX}_{I}^2}{MSE})$ where ${MAX}_I$ is the maximum intensity of the volume and MSE is the mean squared error between volumes $X$ and $\hat{X}$. 

\begin{table}[t]
\centering
\small
\begin{tabular*}{\textwidth}{c|@{\extracolsep{\fill}}cccc}
\Xhline{2\arrayrulewidth}
T2   & REPLICA \cite{RF} & LSDN \cite{LSDN}  & 2D-CNN \cite{MRMR}    & RS-Net (proposed)       \\ \hline
SSMI & 0.901 (0.01)      & 0.909 (0.02)      & 0.929 (0.17)          & \textbf{0.934 (0.02)}   \\ 
PSNR & 28.62 (1.69)      & 30.12 (1.62)      & 30.96 (1.85)          & \textbf{31.13 (1.78)}   \\ \Xhline{2\arrayrulewidth} \\
\Xhline{2\arrayrulewidth}
FLAIR & REPLICA \cite{RF} & LSDN \cite{LSDN}  & 2D-CNN \cite{MRMR}    & RS-Net (proposed)       \\ \hline
SSMI  & 0.870 (0.01)      & 0.887 (0.01)      & 0.897 (0.01)          & \textbf{0.900 (0.01)}   \\ 
PSNR  & 28.32 (1.38)      & 29.68 (1.56)      & 30.32 (1.61)          & \textbf{30.88 (1.84)}   \\ \Xhline{2\arrayrulewidth}  
\end{tabular*}
\caption{Quantitative results for T1-to-T2 (top) and T1-to-FLAIR (bottom) synthesis based on PSNR and SSIM. Higher values indicate better performance. Values in bracket represent standard deviation across volumes. Absolute highest performing results seen in bold. }
\label{brats15}
\end{table}

In order to compare our results to those in the paper \cite{MRMR}, experiments were performed on the 2015 MICCAI BraTS training dataset \cite{brats}. This dataset consists of High-Grade Glioma (HGG) and Low-Grade Glioma (LGG) cases. 54 LGG cases were acquired with T1, T2, T1ce, and FLAIR. Four tumour sub-classes were defined. Volumes are skull-stripped, co-registered, and interpolated to $1 mm^3$ voxel dimension. Each volume is of size 240 x 240 x 155. We follow the same pre-processing steps followed in \cite{MRMR}, where we normalize each volume by dividing by the volume's average intensity. Following \cite{MRMR}, we perform 5-fold cross validation on the dataset (LGG cases). Here, for each cross-validation fold, the dataset is divided into three sets, namely, training, validation, and testing. Each set consists of 42, 6, and 6 volumes respectively. 

\begin{figure*}[t]
\centering
\includegraphics[scale=0.2]{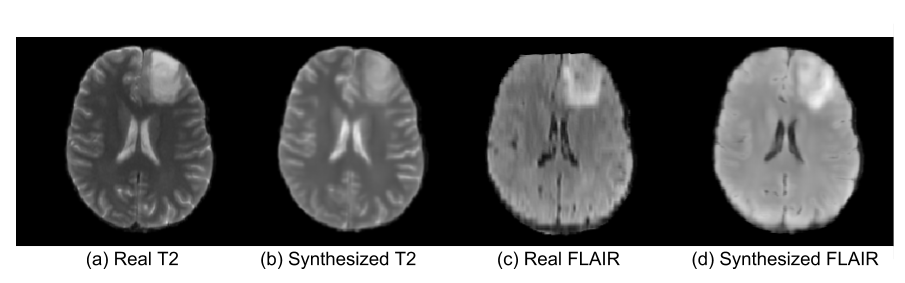}
\center \caption{Example slice from synthetic MR volumes generated by the proposed RS-Net on BraTS 2015 dataset for T1-to-T2 and T1-to-FLAIR synthesis.}
\label{fig:brats-2015}
\end{figure*}

Quantitative comparison of all different methods is given in Table \ref{brats15}. It should be noted that we didn't reproduce the results for other methods and instead report them as listed in \cite{MRMR}. Results indicate that RS-Net performs slightly better than other methods based on the global metrics of PSNR and SSIM, for both T1-to-T2 and T1-to-FLAIR synthesis. The results also show the advantage of using the proposed 3D CNN over 2D CNN. An example showing qualitative results based on RS-Net for both T2 and FLAIR synthesis on a testing volume is shown in Figure \ref{fig:brats-2015}. Note that the resulting MR images are visually similar to the real images, particularly in the area of the tumour.

\subsection{Evaluation of RS-Net synthesis results on downstream tumour segmentation task}
\label{SegEvaluation}
The metrics used in the previous section can be useful in assessing global synthesis quality, but in the context of volumes with pathological structures such as lesions or tumours synthesis quality assessment should focus on the pathological areas. To this end, we quantitatively evaluate the synthesis performance based on their effect on downstream method, tumour segmentation and tumour sub-class segmentation. To this end, we train a new segmentation CNN, for the specific task of multi-class tumor segmentation (referred to as \textbf{S-Net}). This network is similar to the RS-Net but modified such that the synthesis convolution block is removed. S-Net is trained using all 4 real MR volumes with weighted CCE as the loss function. To evaluate the quality of the synthesized volume, one of the real MR volumes is swapped with the synthesized one and the segmentation accuracy is measured. Note that we do \underline{not} retrain the S-Net with the synthesized volume. This allows us to measure quality of the synthesized volumes in comparison to the real volumes.

\subsubsection{Dataset and Pre-processing:}
The 2017 MICCAI BraTS \cite{brats} datasets were used for all the experiments in this section. The BraTS training dataset was used to train the networks. This dataset is comprised of 210 HGG and 75 LGG patients with T1, T1 post contrast (T1ce), T2, and FLAIR MRI for each patient, along with expert tumor labels for each of 3 classes: edema, necrotic/non-enhancing core, and enhancing tumor core. 228 volumes were randomly selected for training the network and another remaining 57 for network validation. A separate BraTS 2017 validation dataset, held out during training, was used to test the synthesis and segmentation performance. This dataset contains 46 patient multi-channel MRI (with no labels provided). The BraTS challenge provided pre-processed volumes that were skull-stripped, co-aligned, and resampled to 1 $mm^3$ voxel volume. The intensities were additionally rescaled using mean subtraction, divided by the standard deviation, and rescaled from 0 to 1 and were cropped to 184 x 200 x 152. For this context, the additional complementary input presented to the regression block (see Figure \ref{fig:SynthSegCNN}(3)) for T1, T2, T1ce, and FLAIR sequences were T1ce, FLAIR, T1, and T2 respectively. This was chosen as T1ce is the gadolinium enhanced version of T1, and FLAIR is the fluid attenuated version of T2.

\begin{figure*}[t]
\centering
\includegraphics[scale=0.22]{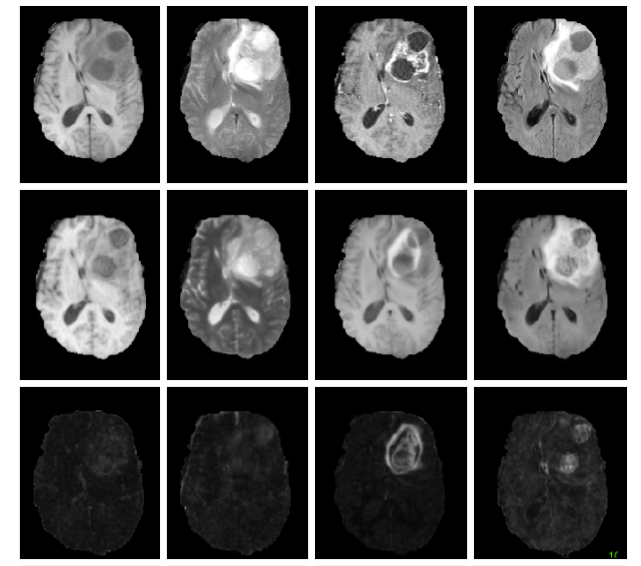}
\center \caption{Example slice from synthetic MR volumes generated using the proposed RS-Net along with its associated uncertainties. Real MRI (Row 1);  synthesized volumes (Row 2) and its associated uncertainty (Row 3) produced as mean and variance across 20 MC dropout samples. Columns from left to right: T1, T2, T1ce, and FLAIR. Notice that uncertainties are highest where predicted tumour enhancements in T1ce are incorrect.}
\label{fig:Uncertainties}
\end{figure*}

\subsubsection{Qualitative Evaluation:}
Synthesis MR volumes produced in a leave-one-out approach by 4 different RS-Nets such that three real MR sequences are used to synthesize the fourth (see Figure \ref{fig:Uncertainties}). The results indicate that the network is able to produce high-quality, high-resolution, 3D synthesized MR volumes, particularly for T1 and T2 sequences, and even for FLAIR. As T1ce shows enhancement within the tumour based on injection of a contrast agent, it was not expected to be easily synthesized from other sequences and error resulted. However, the system indicates locations where the network is uncertain about the regressed output. Qualitative results indicate that errors within the tumour enhancement have associated relatively high uncertainties. This suggests that these uncertainties can be communicated to a clinician or radiologist to indicate trustworthy regions of the synthesized images, and that automatic downstream methods using the synthesized volumes can focus computations on the areas of high confidence, which should be explored in future work.

\begin{table}[t]
\centering
\small
\begin{tabular*}{\textwidth}{c|@{\extracolsep{\fill}}cccc|ccc}
\Xhline{2\arrayrulewidth}
                         & \textbf{T1} & \textbf{T2} & \textbf{FLAIR} & \textbf{T1ce} & \textbf{DE}   & \textbf{DT}   & \textbf{DC}   \\ \Xhline{2\arrayrulewidth}
\textbf{Real}            & \checkmark  & \checkmark  & \checkmark     & \checkmark    & \textbf{68.2} & \textbf{87.9} & \textbf{75.7} \\ \hline 
\textbf{T1 Synthesis}    & $\odot$     & \checkmark  & \checkmark     & \checkmark    & 67.6          & 87.9          & 75.5          \\ \hline
\textbf{T2 Synthesis}    & \checkmark  & $\odot$     & \checkmark     & \checkmark    & 66.3          & 87.3          & 75.6          \\ \hline
\textbf{FLAIR Synthesis} & \checkmark  & \checkmark  & $\odot$        & \checkmark    & 66.8          & 83.6          & 73.1          \\ \hline 
\textbf{T1ce Synthesis}  & \checkmark  & \checkmark  & \checkmark     & $\odot$       & 24.8          & 87.3          & 54.0          \\ \Xhline{2\arrayrulewidth}
\end{tabular*}
\caption{Comparison of multi-class brain tumour segmentation based on S-Net on the BraTS 2017 Validation dataset. The results using all 4 real MRI volumes are compared against replacing 1 real MRI volume with a synthesized MRI volume produced by RS-Net. Notation: Real MR volume (\checkmark), and synthesized MR volume using RS-Net ($\odot$). Quantitative segmentation results based on Dice coefficients for: enhancing tumor (DE), whole tumor (DT), and tumor core (DC).}
\label{tab:Real-Seg}
\end{table}

\subsubsection{Replacing real with synthetic MRI Volumes:}
In Table \ref{tab:Real-Seg}, we compare the tumour segmentation using S-Net in two different testing scenarios, (i) all 4 real MR volumes are provided as input and (ii) 1 real MR volume is replaced with synthesized MR volume for each sequence generated by RS-Net, in turn. We train 4 different RS-Nets to synthesize 4 MR image sequences, where 3 real sequences are presented as input to RS-Net to synthesize the fourth. The synthesized MR volume, along with the 3 real corresponding MR volumes, were then presented to the S-Net previously trained on all four real MRIs. This will allow us to measure quality of the synthesized volume in comparison to the real volume. The resulting labels for BraTS 2017 validation set were uploaded to the BraTS Challenge server, where quantitative segmentation results were provided based on the Dice coefficients for: whole tumor, enhancing tumor, and tumor core. These results (Table \ref{tab:Real-Seg}) indicate that by swapping out real MR volumes with the synthesized T1 or T2 MR volumes generated by the RS-Net leads to comparable brain tumour segmentation performance based on all three reported Dice metrics. For the slightly harder problem of FLAIR synthesis, results indicate a small degradation in tumour segmentation performance for all three Dice metrics. T1ce synthesis results in no loss of whole tumour segmentation performance, but, as predicted, led to a significant reduction in performance in terms of enhancement and necrotic core. This was expected as T1ce is a challenging MRI to synthesize due to its reliance on a contrast agent, which is not used by any other MR sequences.

\begin{table}[t]
\centering
\small
\begin{tabular*}{\textwidth}{c|@{\extracolsep{\fill}}cccc|ccc}
\Xhline{2\arrayrulewidth}
                         & \textbf{T1} & \textbf{T2} & \textbf{FLAIR} & \textbf{T1ce} & \textbf{DE}   & \textbf{DT}   & \textbf{DC}   \\ \Xhline{2\arrayrulewidth}
\textbf{Real}            & \checkmark  & \checkmark  & \checkmark     & \checkmark    & \textbf{68.2} & \textbf{87.9} & \textbf{75.7} \\ \hline 
\textbf{T1 Synthesis}    & $\odot$     & \checkmark  & \checkmark     & \checkmark    & 67.6          & 87.9          & 75.5          \\ 
                         & $\bullet$   & \checkmark  & \checkmark     & \checkmark    & 67.5          & 87.8          & 75.3          \\  \hline
\textbf{T2 Synthesis}    & \checkmark  & $\odot$     & \checkmark     & \checkmark    & 66.3          & 87.3          & 75.6          \\ 
                         & \checkmark  & $\bullet$   & \checkmark     & \checkmark    & 66.1          & 87.2          & 75.4          \\ \hline
\textbf{FLAIR Synthesis} & \checkmark  & \checkmark  & $\odot$        & \checkmark    & 66.8          & 83.6          & 73.1          \\
                         & \checkmark  & \checkmark  & $\bullet$      & \checkmark    & 62.9          & 81.3          & 71.5          \\ \hline 
\textbf{T1ce Synthesis}  & \checkmark  & \checkmark  & \checkmark     & $\odot$       & 24.8          & 87.3          & 54.0          \\ 
                         & \checkmark  & \checkmark  & \checkmark     & $\bullet$     & 24.1          & 85.9          & 53.9          \\ \Xhline{2\arrayrulewidth}
\end{tabular*}
\caption{Comparison of multi-class brain tumour segmentation results based on S-Net on the BraTS 2017 Validation dataset, where each real MR input volume is replaced by its corresponding synthesized MR volume generated by either RS-Net or R-Net in a leave-one-out fashion. Notation: Real MR volume (\checkmark), synthesized MR volume using RS-Net ($\odot$), and R-Net ($\bullet$). Quantitative segmentation results based on Dice coefficients for: enhancing tumor (DE), whole tumor (DT), and tumor core (DC).}
\label{tab:Synth-Seg vs Synth}
\end{table}

\subsubsection{Effectiveness of combined Regression-Segmentation task:}
RS-Net has two output streams for synthesis and segmentation tasks. To check how RS-Net performs in comparison to a network which is trained only for the task of synthesis, we train a new network (\textbf{R-Net}) which is similar to RS-Net but modified such that the segmentation block is removed as well as the additional input to the regression block, and training is based only on weighted MSE. R-Net was trained for the synthesis of all 4 MR image sequences separately, in a leave-one-out approach, and tested for tumor segmentation using S-Net on the BraTS validation dataset exactly as described above. From Table \ref{tab:Synth-Seg vs Synth}, we can observe that R-Net performs comparably to RS-Net, when T1 and T2 are synthesized but shows a small degradation in performance for FLAIR and T1ce synthesis on all three Dice metrics. This shows that performing synthesis and segmentation together allows the network to focus more on tumour part, and in turn gives better quality of the synthesized volume, especially for FLAIR and T1ce.

\begin{table}[t]
\centering
\small
\begin{tabular*}{\textwidth}{c|@{\extracolsep{\fill}}cccc|ccc}
\Xhline{2\arrayrulewidth}
                         & \textbf{T1} & \textbf{T2} & \textbf{FLAIR} & \textbf{T1ce} & \textbf{DE}   & \textbf{DT}   & \textbf{DC}   \\ \Xhline{2\arrayrulewidth}
\textbf{Real}            & \checkmark  & \checkmark  & \checkmark     & \checkmark    & \textbf{68.2} & \textbf{87.9} & \textbf{75.7} \\ \hline 
\textbf{T1 Synthesis}    & $\odot$     & \checkmark  & \checkmark     & \checkmark    & 67.6          & 87.9          & 75.5          \\ 
                         & $\times$    & \checkmark  & \checkmark     & \checkmark    & 66.4          & 85.2          & 71.0          \\  \hline
\textbf{T2 Synthesis}    & \checkmark  & $\odot$     & \checkmark     & \checkmark    & 66.3          & 87.3          & 75.6          \\ 
                         & \checkmark  & $\times$    & \checkmark     & \checkmark    & 66.5          & 87.0          & 71.1          \\ \hline
\textbf{FLAIR Synthesis} & \checkmark  & \checkmark  & $\odot$        & \checkmark    & 66.8          & 83.6          & 73.1          \\
                         & \checkmark  & \checkmark  & $\times$       & \checkmark    & 70.5          & 82.6          & 74.0          \\ \hline 
\textbf{T1ce Synthesis}  & \checkmark  & \checkmark  & \checkmark     & $\odot$       & 24.8          & 87.3          & 54.0          \\ 
                         & \checkmark  & \checkmark  & \checkmark     & $\times$      & 23.1          & 86.5          & 52.0          \\ \Xhline{2\arrayrulewidth}
\end{tabular*}
\caption{Comparison of multi-class brain tumour segmentation results based on S-Net against the results generated directly from the segmentation module of RS-Net for the BraTS 2017 Validation dataset. Notation: Real MR volume (\checkmark), synthesized MR volume using RS-Net ($\odot$), and segmentation output of RS-Net without MR volume ($\times$). Quantitative segmentation results based on Dice coefficients: enhancing tumor (DE), whole tumor (DT), and tumor core (DC).}
\label{tab:Synth-Seg vs Seg}
\end{table}

\subsubsection{Performance of Segmentation part of RS-Net:}
One of the advantages of the RS-Net is that, in addition to MRI synthesis, it also provides tumour segmentation labels. In this section, we will analyze this segmentation part of RS-Net (Figure \ref{fig:SynthSegCNN} (2)). Table \ref{tab:Synth-Seg vs Seg} indicates that the segmentation performance based on RS-Net directly is lower than the results based on using all 4 real MR volumes in S-Net, but is generally lower in comparison to the segmentation results when synthesized MR volumes generated by RS-Net is used in place of a real MR volumes. This trend is consistent across all MR image sequences for all three Dice metrics, except for FLAIR where the enhancing and core tumour Dice is higher for segmentation directly from the RS-Net over the segmentation results from S-Net with a synthesized input (for unknown reasons).

\section{Conclusions}
In this paper, a full resolution 3D end-to-end CNN was developed for the task of MR volume synthesis in the presence of brain tumours. The network was trained for the concurrent tasks of synthesizing a missing MRI sequence and tumour sub-tissue segmentation. Experimental results on BraTS 2015 challenge dataset indicated that the proposed method outperforms all previous methods in terms of traditional evaluation metrics like PSNR and SSIM. The quality of the synthesized images was further evaluated by assessing their effects on the performance in independent tumour segmentation experiments. Experiments on the BraTS 2017 challenge dataset indicated that multi-task learning helps in synthesizing high quality volumes over synthesis alone particularly in more challenging contexts (i.e. FLAIR and T1ce). Results indicated that real MRIs can be replaced with synthesized T1, T2, and FLAIR volumes with minimum degradation in segmentation accuracy, whereas synthesizing T1ce is still too challenging for the task of tumour enhancement segmentation. However, uncertainty measure based on Monte Carlo dropout was shown to be helpful in communicating the confidence in the synthesis results, which will be essential for their adoption by clinicians and downstream automatic methods. The code for the proposed method is available here: \url{https://github.com/RagMeh11/RS-Net}. 

\subsubsection{Acknowledgment}

This work was supported by a Canadian Natural Science and Engineering Research Council (NSERC) Collaborative Research and Development Grant (CRDPJ 505357 - 16) and Synaptive Medical. We gratefully acknowledge the support of NVIDIA Corporation for the donation of the Titan X Pascal GPU used for this research.

\end{document}